\documentclass[accepted]{uai2023} 

\usepackage[american]{babel}

\usepackage{natbib} 
\usepackage{mathtools} 
\usepackage{booktabs} 
\usepackage{tikz} 

\usepackage{amsmath}
\usepackage{amssymb}
\usepackage{mathtools}
\usepackage{amsthm}
\usepackage{multirow}
\usepackage{microtype}
\usepackage{graphicx}
\usepackage{subfigure}
\usepackage{booktabs} 
\usepackage{color}

\title{UAKNN: Label Distribution Learning via Uncertainty-Aware KNN}

\author[1]{Pu Wang}
\author[2]{Yu Zhang}
\author[3]{Zhuoran Zheng}
\affil[1]{%
    the School of Mathematics\\
    Shandong University\\
   Jinan 250100 ,China, 202411943@mail.sdu.edu.cn
}
\affil[2]{
    University of Chinese Academy of Sciences\\
    China, zhengzr@njust.edu.cn
}
\affil[3]{
    the School of cyber science and technology\\
    Sun Yat-sen University\\
    Shenzhen 518000, China, zhengzr@njust.edu.cn
}

\begin{document}
\maketitle

\begin{abstract}
\vspace{-3mm}
Label distribution learning (LDL) aims to characterize the polysemy of an instance by building a set of descriptive degrees corresponding to the instance.
In recent years, researchers seek to model to obtain an accurate label distribution by using low-rank, label relations, expert experiences, and label uncertainty estimation. 
In general, these methods are based on algorithms with parameter learning in a linear (including kernel functions) or deep learning framework.
%
%
However, these methods are difficult to deploy and update online due to high training costs, limited scalability, and outlier sensitivity.
To address this problem, we design a novel LDL method called UAKNN, which has the advantages of the KNN algorithm with the benefits of uncertainty modeling.
In addition, we provide solutions to the dilemma of existing work on extremely label distribution spaces.
Extensive experiments demonstrate that our method is significantly competitive on 12 benchmarks and that the inference speed of the model is well-suited for industrial-level applications.	
 
\end{abstract}

\section{Introduction}\label{sec:intro}
Recently, label distribution learning (LDL)~\cite{geng2016label} gains popularity due to its ability to convey rich semantics for a single instance. 
In contrast to existing multi-label learning paradigms, LDL describes the polysemy of an instance using a set of descriptive degrees to characterize the proportion of each object in an instance.
For example, for an image including \texttt{sun}, \texttt{sky}, and \texttt{tree}, LDL characterizes this instance by the percentage of these three objects (\{10\%, 50\%, 40\%\}) in the image as the descriptive degree, and the sum of the descriptive degrees is 1.
LDL philosophy is successfully introduced to several applications (facial recognition~\cite{chen2020label}, beauty analysis~\cite{ren2017sense}, and age estimation~\cite{gao2017deep}) and achieves significant effectiveness in boosting the generalization capability of the model.
To build a pipeline for LDL, the primary step is to generate a label distribution space.
Up to now, there are two basic approaches for the construction of label distribution spaces: i) Expert-based annotation (manual annotation) approaches, however, these approaches are more subjective and ambiguous, which leads to greater uncertainty in the annotation results; ii) Another solution is the label enhancement algorithm~\cite{xu2019label} which generates the label distribution space by using the features of feature space and the features of logical label space. 
But the label enhancement algorithm lacks theoretical standards, which leads to the generated label distribution space being inaccurate or noisy.
Overall, the currently released label distribution datasets exist with a high probability of inaccuracy and uncertainty, which causes the performance of most LDL algorithms to be limited.

Facing this characteristic of LDL tasks, almost all works attempt to alleviate the uncertainty and inaccuracy of the label distribution space.
On the one hand, modeling a label relationship as a regularization term is enforced on the model during the training stage~\cite{geng2016label,jia2018label,zhao2018label,wang2021label,zheng2018label,jia2021label,li2022unimodal};
On the other hand, creating a higher-order version (distribution of distribution) of the label distribution to estimate the uncertainty of the label space~\cite{zheng2022label}.
Although these methods play a critical role to improve the generalization ability of LDL models, the costly training and the high sensitivity of anomaly information cause challenges in deploying these models.
Specifically, parameterization-based algorithms need to adapt a learner (usually building one or more parameter matrices) for each label distribution dataset, while such models require a well-modulated set of training schemes (including learning rate, number of iterations, and data augmentation techniques, etc.) on different datasets.
When a new batch of the dataset is added to the database, the parameterization-based algorithm requires retraining or fine-tuning, and the model under fine-tuning is highly perturbed by the newly added dataset with anomalous samples.
Based on the strengths and weaknesses of these LDL works, our goal is to build a non-pro parameterized and easy-to-deploy algorithm that is adapted to LDL.
So far, we develop a KNN-type algorithm with uncertainty awareness, named UAKNN, which combines the low-rank characteristic, uncertainty estimation, and robustness (micro-perturbation).
In terms of execution, our algorithm can run on the GPU shader to ensure a real-time response for each test sample inference.
In addition, we propose an ensemble learning strategy and a new evaluation metric (it can be regarded as a loss function) to help existing models learn extremely label distribution datasets.
Some discussions and limitations about UAKNN are also discussed in the paper.
%
\textbf{Our contribution includes:} a) We develop a parameter-free method (UAKNN) for LDL that does not require high training costs and is not sensitive to outliers. 
b) We introduce uncertainty estimation, micro-perturbation, and a scheme with low-rank characteristics to boost the modeling capability of the KNN algorithm.
c) Extensive experiments demonstrate that our algorithm is ideally competitive in terms of inference speed and regression accuracy.

\section{Related work}
This section introduces some works to evaluate the importance of our work, which we have divided into four parts to launch our proposed method.

\noindent \textbf{Label distribution learning.}
LDL serves as a special case of multi-label learning, which characterizes the polysemy of instances with a rich pattern of expressions.
Existing LDL methods~\cite{zheng2022label,geng2016label,jia2018label,jia2021label,ren2017sense,ren2019label,ren2019labelf,zhao2018label,gao2017deep,wang2021label,xu2019label} focus on a paradigm of learning with parameters, including linear models and deep networks.
These learning algorithms with parameters are already significantly competitive on LDL tasks by considering regularization techniques such as low rank, label relations, manifold assumptions, and uncertainty estimation.
However, since numerous forms of label distribution datasets can be constructed, learning methods with parameters require a well-designed system for each dataset.
Such algorithms require high training costs and are sensitive to noisy data.
Fortunately, Geng et al.~\cite{geng2016label}propose the AA-KNN algorithm to clear this trouble, which regresses a label distribution by ``distance'' (p-norm, cosine, etc.) to search for similar samples.
Although KNN-type algorithms~\cite{zhang2007ml,zhang2017efficient,zhang2017learning} can overcome this challenge, KNNs without regularization terms struggle to unlock their potential.

\noindent \textbf{Prototype learning.}%
Learning vector quantization (LVQ)~\cite{kohonen2001learning} is the starting task for prototype learning.
LQV is divided into two main categories: a rule-based scheme~\cite{ren2022prototype} and an optimization of the regularization term or network architecture~\cite{deng2021variational,dong2018few,li2021adaptive}.
Inspired by the rule-based approach, we conduct rule splitting on the training set of the LDL dataset to introduce low-rank characteristics to the KNN-style model.

\noindent \textbf{Uncertainty estimation.}
Uncertainty estimation~\cite{lakshminarayanan2017simple} is broadly employed in tasks such as image recognition~\cite{zhang2021uncertainty}, text classification~\cite{abdar2021review}, and speech recognition~\cite{oneactua2021evaluation}, which boosts the robustness of the model as well as provides interpretation for the algorithm's decisions.
Currently, uncertainty estimation also plays an important role in LDL tasks, which evaluate the inexact and inaccurate label space.
Zheng et al.~\cite{zheng2022label} propose an implicit representation method to build an LDL matrix to evaluate the inaccurate label space within the neural network.
Inspired by this, we expect to impart different weights (obtained by uncertainty sampling) to the searched samples to assemble an accurate label distribution.
This kind of reliance on uncertainty to assign weights also has a large body of work being proposed.

\noindent \textbf{Micro-perturbation scheme.}
Micro-perturbation becomes a universal tool to boost the robustness of the model~\cite{chu2022micro,han2022umix,zhang2021universal,chen2021empirical,hendrycks2019benchmarking}.
Currently, two schemes prove to be successful, one is to enforce noise or synthetic samples on the training targets, and the other is to perturb the parameters of the model directly.
The role of micro-perturbation is to extend the decision boundary of the model or to provide new views for the model.
In our work, micro-perturbations are enforced on the weight generation, which in essence enhances the uncertainty.

\section{Background and Motivation}
Starting in 2016, LDL is officially proposed as a novel learning paradigm that aims to address the polysemy of age estimation~\cite{geng2016label}.
Meanwhile, this work proposes several parameterized learning algorithms and a standard KNN model (AA-KNN) for LDL.
The standard KNN searches for label distributions corresponding to similar samples and afterward estimates the expectation of these label distributions in a simple manner.
Since parameterization-free methods underperform, a large number of researchers focus on parameterized learning algorithms.
In contrast, we propose an assumption: \emph{When does this KNN algorithm with low-rank characteristics, uncertainty estimation capability can break the barrier of this algorithm?}

\textbf{\textcolor{blue}{(a)}} First of all, inspired by prototype learning, to make UAKNN have \textbf{low-rank characteristics}, we divide the training set into several clusters (prototype), each of which is orthogonal to the others, and UAKNN needs to assemble an accurate result for the test samples in these clusters.

\textbf{\textcolor{blue}{(b)}} Secondly, inspired by the \textbf{uncertainty-based estimation} of the Mix-type algorithm, we need to re-weight each of the selected prototypes (samples of the training set).

\textbf{\textcolor{blue}{(c)}} Finally, to obtain a \textbf{smooth decision threshold}, inspired by micro-perturbation learning, the re-assigned weights have a tiny amount of perturbation signal (this signal comes from sampling a distribution).

Compared to existing LDL methods, UAKNN has two unmatched strengths.
No training cost and incremental learnability (the newly added batch of data can be launched directly into the training set).

\section{Method}
In this section, we introduce the technical details of UAKNN.
UAKNN's key philosophy is to search for the closest samples in the ``distance'' and then to assemble them into an accurate result.
We first introduce the basic procedure of UAKNN, which is followed by a theoretical analysis.

\noindent \textbf{Notation.}
Given a particular example, the goal of our method is to learn the degree to which each label describes that instance. 
Input matrix  (tabular data) $\mathcal{X} \in \mathbb{R}^{M \times N}$, where $M$ is the number of instances and $N$ is the dimension of features.
%
%
We define the \textit{i-th} instance in the dataset as $x_{i}$. 
The label distribution space is defined as $\mathcal{Y} \in \mathbb{R}^{M \times L} $, then the \textit{j-th} label is defined as $y_{j}$. 
For each instance $x_{i}$, we define its label distribution $\mathcal{D}_{i}=\left\{{d_{x_{i}}^{y_{1}},d_{x_{i}}^{y_{2}},... ,d_{x_{i}}^{y_{L}} }\right\} $, where $ d_{x_{i}}^{y_{j}} $ is the description degree of the label $y_{j} $ for the instance $x_{i} $. 
The $d_{x_{i}}^{y_{j}}$ is constrained by ${{d_{x_{i}}^{y_{j}}}}\in[0,1] $ and $\sum_{j=1}^{L}{d_{x_{i}}^{y_{L}}}=1$.
In addition, the prototype space ($c$ clusters) is defined as $\mathcal{P} = \{ p_{1}, ...,  p_{c}\}$.
The label distribution that is predicted by the model is defined as $\mathcal{L}_{i} = \left\{{l_{x_{i}}^{y_{1}},l_{x_{i}}^{y_{2}},...,l_{x_{i}}^{y_{L}} }\right\}$.

\noindent \textbf{Vanilla KNN-type method for LDL.}
The vanilla KNN algorithm can be naturally extended to deal with label distribution.
Specifically, given a new instance $x$, its K nearest neighbors are first searched in the training set.
Then, the mean of the label distributions of all the K nearest neighbors is calculated as the label distribution of $x$,
\begin{equation}
\label{eq1}
p\left(y_j \mid x\right)=\frac{1}{k} \sum_{x_{i} \in N_k(x)} d_{x_i}^{y_j},(j=1,2, \ldots, c),
\end{equation}
where $N_k(x)$ is the index of the K nearest neighbors of $x$ in the training set.
From Eq.~\ref{eq1}, it can be seen that K samples are assembled uniformly in a linear manner.
Unfortunately, this ``egalitarianism'' usually does not perform well on real-world datasets due to its not estimating the disturbance of noisy samples.
To this end, we introduce a simple yet effective method, UAKNN, which is based on the philosophy of Eq.~\ref{eq1} to help the model focus on the most similar samples.

\begin{figure}[t]\scriptsize
	\centering
	\includegraphics[width=0.89\linewidth]{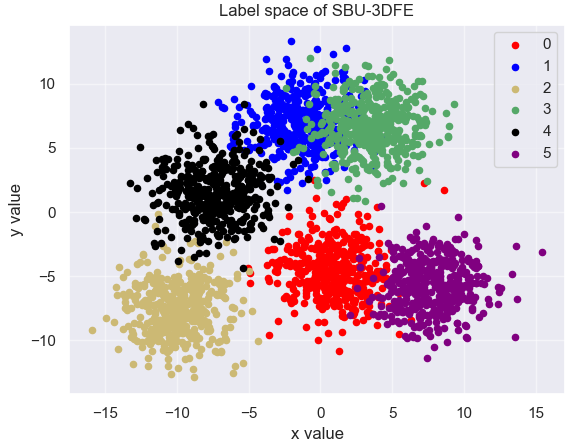}
	\vspace{-4mm}
	\caption{We visualize the label space of the SBU-3DFE dataset by using the t-SNE algorithm~\cite{van2008visualizing}, where t-SNE is based on the KPCA algorithm~\cite{yang2005kpca}. }
	\vspace{-5mm}
	\label{fig-GridMap}
\end{figure}

\noindent \textbf{UAKNN for LDL.}
The low-rank characteristic is employed widely in machine learning algorithms to avoid disturbances caused by noisy samples, where the principle is that a sample can be obtained by the linear combination of several orthogonal samples~\cite{ye2004generalized}.
We introduce this characteristic in the search phase of UAKNN.
Before introducing this characteristic, we observe an interesting property of the label space of the label distribution dataset (see Figure~\ref{fig-GridMap}).
The label space of this dataset (\textit{SBU-3DFE}) has 6 dimensions, which correspond coincidentally to the 6 clusters.
With this, we attempted to reconstruct an accurate label distribution with weights by treating these 6 clusters as 6 basic prototypes (or 6 sets of samples with orthogonal relations).
Specifically, we start with building the prototype space $\mathcal{P}$ on the training dataset $\mathcal{X}$.
$L$ subsets are constructed, and each subset stores the vectors $\mathcal{D}$ that can represent this label.
%
%
%
The formal expression under the Python style:

\begin{equation}
\begin{aligned}
\text{prototype}[\text{i}, :] = \mathcal{D}\underbrace{[\text{np.where}(\mathcal{D}_{i}[\text{i}] > (1/L)), :]}_{\color{blue}{p_{i}}}, \text{i} \in L.
\end{aligned}
\end{equation}

Here, $\mathcal{P}$ includes 6 prototypes (\{$p_{1}, p_{2}, ..., p_{6}$\}), where the sum of the sample numbers of the 6 prototypes is equal to the number of samples in the training set.
In contrast to vanilla KNN, we first estimate the uncertainty of each prototype and then use this quantity to build importance weights (i.e., the higher the uncertainty, the higher the weight, and vice versa).
For the $i$-th label distribution $\mathcal{D}_{i}$ (each prototype provides a sample of the corresponding label distribution), we denote its importance weight as $w_{i}$. Once we obtain the importance weight, we can conduct a weighted linear combination on these $L$ samples by:

\begin{equation}
\mathcal{L} = \text{Softmax}^{*}(\sum_{i=1}^{L}w_{i}\mathcal{D}_{i}),
\end{equation}
where $\mathcal{D}_{i}$ is the label distribution corresponding to the most similar sample obtained by conducting the cosine algorithm in the prototype with the test sample. 
$\text{Softmax}^{*}$ is a Softmax-style normalized operator.
$\mathcal{L}$ denotes the accurate label distribution calculated by UAKNN.

\noindent \textbf{Uncertainty-aware weights.}
So far, we propose how to obtain the weights $w_{i}$ with uncertainty attribute.
First, we assume a reasonable architecture to linearly obtain $w_{i}$ without any hyperparameters,

\begin{equation}
w_{i} = \mathcal{S}(p_{i}) + c_{i},
\end{equation}
where $\mathcal{S}(\cdot)$ is a composite function whose function is to obtain the cosine distance from the closest sample to the test sample in this prototype $p_{i}$, build a Gaussian function, sample, and obtain the mean value of the sampling space. In addition, $c$ (viewed as a micro-perturbation) is a constant value to smooth the obtained results.
Specifically, this strategy involves \textbf{four steps}.

\texttt{Step 1}: Test sample $x_{t}$ searches for the most similar sample in each prototype (cluster) $\{p_{1}, ..., p_{L}\}$ with the help of cosine distance. These $L$ cosine values are denoted as $\{\mu_{1}, ..., \mu_{L}\}$. To make the result of these weights sum to 1, these cosine values are normalized by Softmax.

\texttt{Step 2}: Using these cosine values to construct $L$ Gaussian functions $\{Gf_{1}, ..., Gf_{L}\}$, the value domain of these Gaussian functions is cut off by the Clip function and limited to between [0,1]. The Gaussian function $GF_{i}$ has a mean of $\mu_{i}$, and based on experimental experience, the variance is set to 0.5 to balance sensitivity and robustness.

\texttt{Step 3}: We built $L$ sets $\{T_{1}, ..., T_{L}\}$, and the elements in each set $T_{i}$ are the result of a random sampling of $GF_{i}$. Each set $T_{i}$  has 100 elements. $\mathcal{S}(p_{i})$ is obtained by calculating the mean value for each set $T_{i}$.

\texttt{Step 4}: To introduce minor perturbations and control the output scale for obtaining 
$c_{i}$, construct a new set $C$ where each element is set to 
5\% of $\mu_{i}$. Using these as the mean of the Gaussian functions, repeat \texttt{Steps 2-3}.

\noindent \textbf{Theories.}
We theorize the generalized upper bound of the whole model easily.
Note that the cosine estimate is treated as the probability of correct classification.
Fundamentally, we address the $L-\textit{NN}$ problem with the following theoretical derivation.
Given the test sample $x_{t}$, and the nearest neighbor sample $z$, the probability of error is:
\begin{equation}
P(error) = 1 - \sum_{c \in \mathcal{P}, c \leq L}P(c|x)P(c|z).
\end{equation}
Assuming that the samples are \textit{i.i.d.} then $b^{*}$ denotes the bayesian optimal decision maker,
\begin{equation}
P(error) = L^{2} \times (1-P(b^{*}|x_{t})).
\end{equation}
The generalization error rate of $L-\textit{NN}$  is less than the error rate of $L^{2}$ $\times$ the bayesian optimal decision maker.

\noindent \textbf{Discussions.}
For the elaborated pipeline, we are required to explore the reasonableness and usefulness of the algorithm. We divide \textbf{\textcolor{red}{five}} sub-issues to illustrate UAKNN.

1) \noindent \textit{Why are uncertainty-aware weights workable?}
Theoretically, KNN is an empirical risk minimization (ERM) method.
ERM-type methods are designed with the risk of overfitting when having an untrustworthy training dataset.
Uncertainty awareness or estimation enforced on the algorithm inherently expands the decision boundary.
We evaluate the entropy values of the label space obtained from vanilla KNN and UAKNN performs on the SBU-3DFE dataset, and UAKNN boosts by 6\% compared to vanilla KNN.
In the experimental section, we set a baseline (WUAKNN) to illustrate the effectiveness of the uncertainty.

2) \noindent \textit{The label distribution is standardized (linear or Softmax).}
The output morphology (the range of values is [0, 1] and the sum is 1) of the label distribution is defined in a ``textbook'' fashion.
Almost all the outputs of the label distribution algorithm reach the definition of label distribution with the help of Softmax.
We retain the structure of Softmax but replace the natural exponent base \textbf{e} with \textbf{2} ($\text{Softmax}^{*}$), primarily to address the issue where the nonlinear amplification by $e^{x}$ overly exaggerates the gaps between labels, leading to overconfident probabilities clustered near 0 or 1. By using 
$2^{x}$, which grows slower than $e^{x}$, we achieve a more controlled and moderate amplification of label differences. This adjustment balances sensitivity and stability, aligning with tasks requiring calibrated uncertainty, such as multi-label ambiguity or medical diagnosis.

3) \noindent \textit{Overcome false outcomes.}
In the LDL task, there is an interesting phenomenon that the results are usually very competitive when the predicted label space tends to be uniformly distributed.
This is counter to the fact since LDL evaluation systems usually consider the ``distance'' between vectors.
UAKNN tackles this challenge thanks to low-rank characteristics and uncertainty and micro-perturbation strategies.
%
We count the variance (average value) of the predicted label distribution on 12 benchmarks for UAKNN (0.162) and WUAKNN (0.148) respectively, and UAKNN has a 12\% higher variance than WUAKNN.

4) \noindent \textit{Inference speed of the model.}
Since our algorithm involves a lot of sampling operations ($\mathcal{S}(\cdot)$, and $c$), we design a pipeline to obtain real-time inference capability.
First, we quantize the UAKNN data in the PyTorch 2.0 framework and encapsulate it into a pipeline by \texttt{skorch}\footnote[1]{\textcolor{blue}{\url{https://github.com/skorch-dev/skorch}}}.
Finally, we compile this pipeline with the help of \texttt{torch.compile}\footnote[2]{\textcolor{blue}{\url{https://pytorch.org/get-started/pytorch-2.0/}}}.
Our algorithm achieves real-time inference on all 12 public datasets (120 test samples / second on the GPU shader).
Note that UAKNN is optimized by the KDtree~\cite{zhou2008real} and KDball~\cite{zhou2008real} algorithms to accelerate the retrieval speed when facing a dataset with numerous features (\textit{Movie} and \textit{SCUT}).

5) \noindent \textit{The parameter setting of UAKNN is fixed.}
In the description of the algorithm's steps, \texttt{Step 2-4} have artificially set parameters, which are obtained for the analysis of parameter sensitivity.

\section{Experiments}
This section evaluates UAKNN on 12 benchmarks and investigates its parameter sensitivity and effectiveness.

\noindent \textbf{Algorithm configurations.}
We conduct experiments on 12 datasets (including image, text, and tabular formats), and the characteristics of the datasets are summarized in Table~\ref{T1}.
Among them, the \texttt{ID-1} dataset is referenced to Zheng et al.~\cite{zheng2022label} and the rest to Gao et al.~\cite{gao2017deep}.
To evaluate the performance of LDL models, we use the six metrics proposed by~\cite{geng2016label}, including Chebyshev distance $\downarrow$, Clark distance $\downarrow$, Canberra distance $\downarrow$, KL divergence $\downarrow$, Cosine similarity $\uparrow$, and Intersection similarity $\uparrow$.
%
%
$\downarrow$ represents the indicator's performance favoring \textbf{low} values and $\uparrow$ represents the indicator's performance favoring \textbf{high} values.

\noindent \textbf{Experimental setting.}
We conduct comparative experiments with seven LDL algorithms (WUAKNN, INP~\cite{zheng2022label}, BFGS-LLD~\cite{geng2016label}, LDL-LRR ~\cite{jia2021label}, LDL-LCLR~\cite{ren2019label}, LDLSF~\cite{ren2019labelf} and LALOT~\cite{zhao2018label}) on 12 benchmarks.
To evaluate the effectiveness of our framework, we set a baseline (WUAKNN: without uncertainty-aware KNN) as one of the comparison methods, which maintains only prototype matching on the base of UAKNN, and the weights are normalized with $\text{Softmax}^{*}$.
INP proposes an implicit representation to estimate the uncertainty of the label space, which involves designing 12 different deep networks for 12 benchmarks.
BFGS-LLD is based on a linear model, the loss function is K-L divergence, and the quasi-Newton algorithm's optimization scheme.
LDL-LRR and LDL-LCLR both consider label correlations in the learning process, with the former considering the order relationship of the labels and the latter capturing global relationships between labels.
For LDL-LRR, the parameters $\lambda$ and $\beta$ are selected from $10^{\{-6,-5, \ldots,-2,-1\}}$ and $10^{\{-3,-2, \ldots, 1,2\}}$, respectively. For LDL-LCLR, the parameters $\lambda_{1}, \lambda_{2}, \lambda_{3}, \lambda_{4}$ and $k$ are set to $0.0001,0.001,0.001,0.001$ and $4$, respectively.
LDLSF leverages label-specific features and common features simultaneously, whose parameters $\lambda_{1}, \lambda_{2}$ and $\lambda_{3}$ are selected from $10^{\{-6,-5, \ldots,-2,-1\}}$, respectively, and $\rho$ is set to $10^{-3}$. 
LALOT adopts optimal transport distance as the loss function, and the trade-off parameter $C$ and the regularization coefficient $\lambda$ are set to $200$ and $0.2$, respectively.
%
%

\begin{table}[t] \footnotesize
	\begin{center}
		\vspace{-0mm}
		\caption{Statistics of the experimental datasets. These datasets are translated (adopting pre-trained models, such as ResNet18 or BERT) into tabular datasets from images, text, and other patterns.}
		\vspace{-2mm}
		\label{T1}
		\begin{tabular}{llccc}
			\toprule
			ID & Dataset 	          & Examples	& Features  & Labels            \\ 
			\midrule
			1  &  wc-LDL   &  500        & 243       & 12             \\
			2  &  SJAFFE          &  213        & 243       & 6        \\
			3  &  SBU-3DFE        &  2500       & 243       & 6         \\
			4  &  Scene           &  2000       & 294       & 9       \\
			5  &  Gene            &  17892      & 36        & 68      \\
			6  &  Movie           &  7755       & 1869      & 5      \\
			7  &  M2B             &  1240       & 250       & 5        \\
			8  &  SCUT            &  1500       & 300       & 5       \\
			10  &  RAF-ML         &  4908       & 200       & 6       \\
			11  &  Twitter        &  10040      & 200       & 8     \\
			12  &  Flickr         &  11150      & 200       & 8       \\
			\bottomrule
		\end{tabular}%
		\vspace{-6mm}
	\end{center}
\end{table}

\begin{table*}[!htb] \tiny
	\begin{center}
		\vspace{-0mm}
		\caption{The performance of our proposed method with the comparison algorithms on 12 datasets. All algorithms are run on an RTX3090 GPU shader. }
		\vspace{-2mm}
		\label{T3}
		\resizebox{\linewidth}{!}{
			\begin{tabular}{c|c|cccccc}
				\toprule
				Dataset                             & Algorithm	   & Chebyshev $\downarrow$  & Clark $\downarrow$   &Canberra $\downarrow$   &K-L $\downarrow$    &Cosine $\uparrow$   &Intersection $\uparrow$           \\ 	
				\midrule 
				
				&      Ours     & 0.0749 $\pm$ \text{0.0015}           & 0.3899 $\pm$ \text{0.0051}       &0.7679 $\pm$ \text{0.0035}        & 0.04021 $\pm$ \text{0.0008}     & 0.9899 $\pm$ \text{0.0013}      & 0.8819 $\pm$ \text{0.0015}                \\
				
				&     WUAKNN     & 0.0788 $\pm$ \text{0.0021}           & 0.4013 $\pm$ \text{0.0042}       &0.7772 $\pm$ \text{0.0031}          & 0.04093 $\pm$ \text{0.0056}     & 0.9813 $\pm$ \text{0.0015}      & 0.8761 $\pm$ \text{0.0019}                \\
				
				&     INP     & 0.0779 $\pm$ \text{ 0.0021}           & 0.3980 $\pm$ \text{0.0051}       &0.7779 $\pm$ \text{0.0030}          & 0.04040 $\pm$ \text{0.0020}     & 0.9883 $\pm$ \text{0.0009}      & 0.8778 $\pm$ \text{0.0014}                \\
				
				&  LDL-LRR     & 0.1122 $\pm$ \text{0.0030}          & 0.4772 $\pm$ \text{0.0036}      &0.8802 $\pm$ \text{0.0024}          & 0.05533 $\pm$ \text{0.0049}     & 0.9510 $\pm$ \text{0.0022}      & 0.8555 $\pm$ \text{0.0047}                   \\
				
				&  LDL-LCLR    & 0.1057 $\pm$ \text{ 0.0019}           & 1.0569 $\pm$ \text{ 0.0039}       &0.7890 $\pm$ \text{ 0.0039}          & 0.05045 $\pm$ \text{ 0.0037}     & 0.9668 $\pm$ \text{ 0.0049}      & 0.8383 $\pm$ \text{ 0.0018}                     \\
				
				&  LDLSF       & 0.1009 $\pm$ \text{0.0038}           & 0.4199 $\pm$ \text{0.0044}       &0.9008 $\pm$ \text{0.0015}          & 0.05199 $\pm$ \text{0.0040}     & 0.9779 $\pm$ \text{0.0018}      & 0.8660 $\pm$ \text{0.0022}                     \\
				
				&  LALOT       & 0.0989 $\pm$ \text{0.0019}          & 0.6689 $\pm$ \text{0.0019}       &0.8089 $\pm$ \text{0.0049}          & 0.04778 $\pm$ \text{0.0018}     & 0.9476 $\pm$ \text{0.0020}      & 0.8700 $\pm$ \text{0.0033}                    \\
				
				\multirow{-7}{*}{wc-LDL} 		            &  BFGS-LLD    & 0.1229 $\pm$ \text{0.0039}           & 1.5657 $\pm$ \text{0.0021}       &0.7998 $\pm$ \text{0.0020}          & 0.04998 $\pm$ \text{0.0051}     & 0.9704 $\pm$ \text{0.0036}      & 0.8611 $\pm$ \text{0.0016}                 \\
				
				\midrule
				
				&  Ours        & 0.0825 $\pm$ \text{0.0025}           & 0.4011 $\pm$ \text{0.0036}       &0.7892 $\pm$ \text{0.0049}          & 0.04015 $\pm$ \text{0.0014}      & 0.9890 $\pm$ \text{0.0034}      & 0.8849 $\pm$ \text{0.0054}                   \\
				
				&  WUAKNN     & 0.0899 $\pm$ \text{0.0035}           & 0.4129 $\pm$ \text{0.0029}       &0.8013 $\pm$ \text{0.0035}          & 0.04224 $\pm$ \text{0.0066}     & 0.9655 $\pm$ \text{0.0014}      & 0.8589 $\pm$ \text{0.0014}                \\
				
				&  INP        & 0.0854 $\pm$ \text{0.0018}           & 0.4008 $\pm$ \text{0.0030}       &0.7955 $\pm$ \text{0.0023}          & 0.04100 $\pm$ \text{0.0012}      & 0.9799 $\pm$ \text{0.0014}      & 0.8809 $\pm$ \text{0.0015}                   \\
				
				&  LDL-LRR     & 0.1122 $\pm$ \text{0.0030}           & 0.4772 $\pm$ \text{0.0036}       &0.8802 $\pm$ \text{0.0024}          & 0.5533 $\pm$ \text{0.0049}     & 0.9510 $\pm$ \text{0.0022}      & 0.8555 $\pm$ \text{0.0047}                       \\
				
				&  LDL-LCLR    & 0.1057 $\pm$ \text{0.0019}           & 1.0569 $\pm$ \text{0.0039}       &0.7890 $\pm$ \text{0.0039}          & 0.5045 $\pm$ \text{0.0037}     & 0.9668 $\pm$ \text{0.0049}      & 0.8383 $\pm$ \text{0.0018}                      \\
				
				&  LDLSF       & 0.1123 $\pm$ \text{0.0038}           & 0.4397 $\pm$ \text{0.0044}       &0.9212 $\pm$ \text{0.0015}          & 0.5557 $\pm$ \text{0.0040}     & 0.9779 $\pm$ \text{0.0018}      & 0.8660 $\pm$ \text{0.0022}                       \\
				
				&  LALOT       & 0.0979 $\pm$ \text{0.0018}           & 0.6799 $\pm$ \text{0.0021}       &0.8077 $\pm$ \text{0.0039}          & 0.4756 $\pm$ \text{0.0015}     & 0.9433 $\pm$ \text{0.0111}      & 0.8423 $\pm$ \text{0.0034}                      \\
				
				\multirow{-6}{*}{SJAFFE} 		            &  BFGS-LLD    & 0.1334 $\pm$ \text{0.0139}           & 1.6648 $\pm$ \text{0.0023}       &0.7999 $\pm$ \text{0.0022}          & 0.0477 $\pm$ \text{0.0051}     & 0.9711 $\pm$ \text{0.0036}      & 0.8655 $\pm$ \text{0.0116}                                        \\
				
				\midrule
				
				&  Ours        & 0.0811 $\pm$ \text{0.0023}           & 0.3987 $\pm$ \text{0.0024}       &0.7533 $\pm$ \text{0.0022}          & 0.03541 $\pm$ \text{0.0033}      & 0.9888 $\pm$ \text{0.0066}      & 0.8997 $\pm$ \text{0.0033}                    \\
				
				&  WUAKNN     & 0.0970 $\pm$ \text{ 0.044}           & 0.4151 $\pm$ \text{0.0088}       &0.7810 $\pm$ \text{0.0023}          & 0.04140 $\pm$ \text{0.0019}     & 0.9711 $\pm$ \text{0.0013}      & 0.8797 $\pm$ \text{0.0016}                \\
				
				&  INP        & 0.0833 $\pm$ \text{0.0020}           & 0.3994 $\pm$ \text{0.0010}       &0.7611 $\pm$ \text{0.0020}          & 0.03650 $\pm$ \text{0.0014}      & 0.9811 $\pm$ \text{0.0015}      & 0.8900 $\pm$ \text{0.0017}                    \\
				
				&  LDL-LRR     & 0.1109 $\pm$ \text{0.0036}           & 0.4477 $\pm$ \text{0.0039}      &0.8666 $\pm$ \text{0.0026}          & 0.05344 $\pm$ \text{0.0028}     & 0.9597 $\pm$ \text{0.0029}      & 0.8592 $\pm$ \text{0.0033}                    \\
				
				&  LDL-LCLR    & 0.1100 $\pm$ \text{0.0025}           & 0.9660 $\pm$ \text{0.0039}       &0.7897 $\pm$ \text{0.0033}          & 0.05101 $\pm$ \text{0.0021}     & 0.9677 $\pm$ \text{0.0056}      & 0.8555 $\pm$ \text{0.0032}                   \\
				
				&  LDLSF       & 0.1009 $\pm$ \text{0.0038}           & 0.4199 $\pm$ \text{0.0044}       &0.9008 $\pm$ \text{0.0015}          & 0.05199 $\pm$ \text{0.0040}     & 0.9780 $\pm$ \text{0.0029}      & 0.8660 $\pm$ \text{0.0022}                     \\
				
				&  LALOT       & 0.0899 $\pm$ \text{0.0021}           & 0.6563 $\pm$ \text{0.0019}       &0.8132 $\pm$ \text{0.0100}          & 0.04668 $\pm$ \text{0.0021}     & 0.9441 $\pm$ \text{0.0011}      & 0.8723 $\pm$ \text{0.0034}                    \\
				
				\multirow{-7}{*}{SBU} 		            &  BFGS-LLD    & 0.1119 $\pm$ \text{0.0030}           & 1.4657 $\pm$ \text{0.0022}       &0.7700 $\pm$ \text{0.0025}          & 0.04932 $\pm$ \text{0.0053}     & 0.9753 $\pm$ \text{0.0036}      & 0.8710 $\pm$ \text{0.0019}                    \\
				
				\midrule
				
				&  Ours        & 0.2993 $\pm$ \text{0.0041}           & 2.3079 $\pm$ \text{0.0089}       &6.4135 $\pm$ \text{0.0031}          & 0.8034 $\pm$ \text{0.0022}      & 0.7995 $\pm$ \text{0.0077}      & 0.5703 $\pm$ \text{0.0003}                    \\
				
				&  WUAKNN     & 0.3146 $\pm$ \text{ 0.0024}           & 2.3550 $\pm$ \text{0.0156}       &6.6990 $\pm$ \text{0.1933}          & 0.8634 $\pm$ \text{0.01006}     & 0.7655 $\pm$ \text{0.0013}      & 0.5322 $\pm$ \text{0.0016}                \\
				
				&  INP        & 0.2998 $\pm$ \text{0.0020}           & 2.3374 $\pm$ \text{0.0018}       &6.5163 $\pm$ \text{0.0018}          & 0.8111 $\pm$ \text{0.0029}      & 0.7890 $\pm$ \text{0.0049}      & 0.5691 $\pm$ \text{0.0010}                    \\
				
				&  LDL-LRR     & 0.3889 $\pm$ \text{0.0111}           & 3.1698 $\pm$ \text{0.0031}      &6.8777 $\pm$ \text{0.0025}          & 0.8999 $\pm$ \text{0.0069}     & 0.7044 $\pm$ \text{0.0077}      & 0.5444 $\pm$ \text{0.0049}                  \\
				
				&  LDL-LCLR    & 0.3740 $\pm$ \text{0.0066}           & 2.4986 $\pm$ \text{0.0066}       &6.8600 $\pm$ \text{0.0067}          & 0.8559 $\pm$ \text{0.0039}     & 0.7119 $\pm$ \text{0.0122}      & 0.5119 $\pm$ \text{0.0081}                   \\
				
				&  LDLSF       & 0.3441 $\pm$ \text{0.0249}           & 2.9884 $\pm$ \text{0.0055}       &6.6900 $\pm$ \text{0.0055}          & 0.8391 $\pm$ \text{0.0044}     & 0.7336 $\pm$ \text{0.0088}      & 0.5660 $\pm$ \text{0.0041}                     \\
				
				&  LALOT       & 0.3129 $\pm$ \text{0.0152}           & 2.3999 $\pm$ \text{0.0044}       &6.6666 $\pm$ \text{0.0078}          & 0.8226 $\pm$ \text{0.0033}     & 0.7390 $\pm$ \text{0.0100}      & 0.5224 $\pm$ \text{0.0066}                   \\
				
				\multirow{-7}{*}{Scene} 		            &  BFGS-LLD    & 0.3598 $\pm$ \text{0.0020}           & 2.4998 $\pm$ \text{0.0033}       &6.7999 $\pm$ \text{0.0049}          & 0.8400 $\pm$ \text{0.0033}     & 0.7333 $\pm$ \text{0.0064}      & 0.5199 $\pm$ \text{0.0055}                    \\
				
				\midrule
				
				&  Ours        & 0.0481 $\pm$ \text{0.0036}           & 2.1010 $\pm$ \text{0.0256}       &14.0802 $\pm$ \text{0.0154}          & 0.2333 $\pm$ \text{0.0091}      & 0.8409 $\pm$ \text{0.0030}      & 0.7993 $\pm$ \text{0.0022}                   \\
				
				&  WUAKNN     & 0.0513 $\pm$ \text{0.0069}           & 2.2019 $\pm$ \text{0.0055}       &14.1489 $\pm$ \text{0.2011}          & 0.2441 $\pm$ \text{0.0122}     & 0.8349 $\pm$ \text{0.0013}      & 0.7829 $\pm$ \text{0.0018}                \\
				
				&  INP        & 0.0488 $\pm$ \text{0.0012}           & 2.1029 $\pm$ \text{0.0259}       &14.0888 $\pm$ \text{0.0551}          & 0.2335 $\pm$ \text{0.0044}      & 0.8395 $\pm$ \text{0.0032}      & 0.7984 $\pm$ \text{0.0066}                   \\
				
				&  LDL-LRR     & 0.0537 $\pm$ \text{0.0039}           & 2.2887 $\pm$ \text{0.0860}      &14.3550 $\pm$ \text{0.0144}          & 0.2559 $\pm$ \text{0.0077}     & 0.8288 $\pm$ \text{0.0144}      & 0.7789 $\pm$ \text{0.0040}                     \\
				
				&  LDL-LCLR    & 0.0511 $\pm$ \text{0.0022}           & 2.2201 $\pm$ \text{0.0444}       &14.2101 $\pm$ \text{0.0510}          & 0.2566 $\pm$ \text{0.0047}     & 0.8302 $\pm$ \text{0.0012}      & 0.7722 $\pm$ \text{0.0060}                    \\
				
				&  LDLSF       & 0.0513 $\pm$ \text{0.0030}           & 2.2221$\pm$ \text{0.0036}       &14.3667 $\pm$ \text{0.0265}          & 0.2445 $\pm$ \text{0.0077}     & 0.8320 $\pm$ \text{0.0010}      & 0.7701 $\pm$ \text{0.0026}                      \\
				
				&  LALOT       & 0.0505 $\pm$ \text{0.0033}           & 2.1989 $\pm$ \text{0.0194}       &14.1855 $\pm$ \text{0.0922}          & 0.2443 $\pm$ \text{0.0088}     & 0.8297 $\pm$ \text{0.0060}      & 0.7888 $\pm$ \text{0.0013}                     \\
				
				\multirow{-7}{*}{Gene} 		            &  BFGS-LLD    & 0.0578 $\pm$ \text{0.0066}           & 2.3008 $\pm$ \text{0.0188}       &14.3559 $\pm$ \text{0.1556}          & 0.2480 $\pm$ \text{0.0015}     & 0.8300$\pm$ \text{0.0049}      & 0.7786 $\pm$ \text{0.0070}                     \\
				
				\midrule
				
				&  Ours        & 0.1077 $\pm$ \text{0.0018}           & 0.4991 $\pm$ \text{0.0035}       &0.9712 $\pm$ \text{0.0040}          & 0.0977 $\pm$ \text{0.0013}      & 0.9582 $\pm$ \text{0.0166}      & 0.8722 $\pm$ \text{0.0015}                     \\
				
				& WUAKNN     & 0.1116 $\pm$ \text{ 0.0044}           & 0.5229 $\pm$ \text{0.0150}       &1.0896 $\pm$ \text{0.0119}          & 0.1366 $\pm$ \text{0.0011}     & 0.9413 $\pm$ \text{0.0345}      & 0.8745 $\pm$ \text{0.0088}                \\
				
				&  INP        & 0.1089 $\pm$ \text{0.0018}           & 0.5001 $\pm$ \text{0.0044}       &0.9722 $\pm$ \text{0.0040}          & 0.0977 $\pm$ \text{0.0008}      & 0.9585 $\pm$ \text{0.0061}      & 0.8861 $\pm$ \text{0.0006}                     \\
				
				&  LDL-LRR     & 0.1135 $\pm$ \text{0.0009}           & 0.5244$\pm$ \text{0.0010}      &1.1551 $\pm$ \text{0.0061}          & 0.1445 $\pm$ \text{0.0049}     & 0.9510 $\pm$ \text{0.0022}      & 0.8772 $\pm$ \text{0.0007}                    \\
				
				&  LDL-LCLR    & 0.1177 $\pm$ \text{0.0086}           & 0.5345 $\pm$ \text{0.0040}       &1.1533 $\pm$ \text{0.0111}          & 0.1559 $\pm$ \text{0.0030}     & 0.9360 $\pm$ \text{0.0049}      & 0.8222 $\pm$ \text{0.0011}                    \\
				
				&  LDLSF       & 0.1155 $\pm$ \text{0.0045}           & 0.5339 $\pm$ \text{0.0062}       &1.1152$\pm$ \text{0.0050}          & 0.1540 $\pm$ \text{0.0041}     & 0.9445 $\pm$ \text{0.0020}      & 0.8551 $\pm$ \text{0.0044}                     \\
				
				&  LALOT       & 0.1221 $\pm$ \text{0.0110}           & 0.5440 $\pm$ \text{0.0033}       &1.111 $\pm$ \text{0.0040}          & 0.1503 $\pm$ \text{0.0008}     & 0.9477 $\pm$ \text{0.0022}      & 0.8559 $\pm$ \text{0.0002}                    \\
				
				\multirow{-7}{*}{Movie} 		            &  BFGS-LLD    & 0.1310 $\pm$ \text{0.0032}           & 0.5230 $\pm$ \text{0.0022}       &1.1170 $\pm$ \text{0.0024}          & 0.1595 $\pm$ \text{0.0155}     & 0.9400 $\pm$ \text{0.0003}      & 0.8491 $\pm$ \text{0.0018}                    \\
				
				\midrule
				
				&  Ours        & 0.3694 $\pm$ \text{0.0025}           & 1.1555 $\pm$ \text{0.0103}       &2.0882 $\pm$ \text{0.0088}          & 0.4883 $\pm$ \text{0.0006}      & 0.8026 $\pm$ \text{0.0034}      & 0.6801 $\pm$ \text{0.0088}                   \\
				
				& WUAKNN     & 0.4004 $\pm$ \text{ 0.0063}           & 1.2899 $\pm$ \text{0.0112}       &2.1998 $\pm$ \text{0.1088}          & 0.5012 $\pm$ \text{0.0045}     & 0.7889 $\pm$ \text{0.0099}      & 0.6521 $\pm$ \text{0.0009}                \\
				
				&  INP        & 0.3763 $\pm$ \text{0.0022}           & 1.1560 $\pm$ \text{0.0102}       &2.0889 $\pm$ \text{0.0055}          & 0.4880 $\pm$ \text{0.0023}      & 0.7998 $\pm$ \text{0.0022}      & 0.6703 $\pm$ \text{0.0033}                   \\
				
				&  LDL-LRR     & 0.3993 $\pm$ \text{0.0010}           & 1.4990 $\pm$ \text{0.0166}      &2.1884 $\pm$ \text{0.0034}          & 0.5246 $\pm$ \text{0.0006}     & 0.7531 $\pm$ \text{0.0023}      & 0.6334 $\pm$ \text{0.0077}                     \\
				
				&  LDL-LCLR    & 0.4040 $\pm$ \text{0.0082}           & 1.2444 $\pm$ \text{0.0045}       &2.2000 $\pm$ \text{0.0009}          & 0.4996 $\pm$ \text{0.0013}     & 0.7760 $\pm$ \text{0.0079}      & 0.6555 $\pm$ \text{0.0012}                 \\
				
				&  LDLSF       & 0.4159 $\pm$ \text{0.0055}          & 1.3105 $\pm$ \text{0.0041}       &2.2155 $\pm$ \text{0.0076}          & 0.5002 $\pm$ \text{0.0006}     & 0.7552 $\pm$ \text{0.0004}      & 0.6234 $\pm$ \text{0.0033}                    \\
				
				&  LALOT       & 0.3881 $\pm$ \text{0.0099}           & 1.4883 $\pm$ \text{0.0012}       &2.1257 $\pm$ \text{0.0268}          & 0.4990 $\pm$ \text{0.0008}     & 0.7549 $\pm$ \text{0.0021}      & 0.6620 $\pm$ \text{0.0053}                    \\
				
				\multirow{-7}{*}{M2B} 		            &  BFGS-LLD    & 0.3811 $\pm$ \text{0.0044}          & 1.3650 $\pm$ \text{0.0002}       &2.1992 $\pm$ \text{0.0095}          & 0.4995 $\pm$ \text{0.0005}     & 0.7699 $\pm$ \text{0.0040}      & 0.6532 $\pm$ \text{0.0009}                   \\
				
				\midrule
				
				&  Ours        & 0.3877 $\pm$ \text{0.0073}           & 1.2597 $\pm$ \text{0.0123}       &2.1925 $\pm$ \text{0.0050}          & 0.4912 $\pm$ \text{0.0009}      & 0.7010 $\pm$ \text{0.0022}      & 0.6946 $\pm$ \text{0.0039}                   \\
				
				&  WUAKNN     & 0.4011 $\pm$ \text{0.0099}           & 1.3461 $\pm$ \text{0.0122}       &2.2119 $\pm$ \text{0.0398}          & 0.5125 $\pm$ \text{0.0088}     & 0.6765 $\pm$ \text{0.0010}      & 0.6402 $\pm$ \text{0.0022}                \\
				
				&  INP        & 0.3895 $\pm$ \text{0.0021}           & 1.2640 $\pm$ \text{0.0111}       &2.1995 $\pm$ \text{0.0095}          & 0.4911 $\pm$ \text{0.0030}      & 0.6990 $\pm$ \text{0.0002}      & 0.6904 $\pm$ \text{0.0001}                   \\
				
				&  LDL-LRR     & 0.4159 $\pm$ \text{0.0010}           & 1.6680 $\pm$ \text{0.0122}      &2.2006 $\pm$ \text{0.0039}          & 0.5388 $\pm$ \text{0.0006}     & 0.6531 $\pm$ \text{0.0023}      & 0.5804 $\pm$ \text{0.0007}                  \\
				
				&  LDL-LCLR    & 0.4240 $\pm$ \text{0.0042}           & 1.3444 $\pm$ \text{0.0055}       &2.2450$\pm$ \text{0.0016}          & 0.5131 $\pm$ \text{0.0022}     & 0.6261 $\pm$ \text{0.0005}      & 0.5500 $\pm$ \text{0.0012}                  \\
				
				&  LDLSF       & 0.4360 $\pm$ \text{0.0015}           & 1.2185 $\pm$ \text{0.0022}       &2.2159 $\pm$ \text{0.0076}          & 0.5120 $\pm$ \text{0.0006}     & 0.6261 $\pm$ \text{0.0004}      & 0.5534 $\pm$ \text{0.0030}                     \\
				
				&  LALOT       & 0.3999 $\pm$ \text{0.0009}           & 1.4983 $\pm$ \text{0.0012}       &2.2207 $\pm$ \text{0.0158}          & 0.4995 $\pm$ \text{0.0002}     & 0.6549 $\pm$ \text{0.0020}      & 0.6411 $\pm$ \text{0.0044}                    \\
				
				\multirow{-7}{*}{SCUT} 		            &  BFGS-LLD    & 0.3992 $\pm$ \text{0.0055}          & 1.5656 $\pm$ \text{0.0163}       &2.2832 $\pm$ \text{0.0080}          & 0.4966 $\pm$ \text{0.0011}     & 0.6491 $\pm$ \text{0.0040}      & 0.6333 $\pm$ \text{0.0013}                    \\	
				\bottomrule
			\end{tabular}%
		}
		\vspace{-6mm}
	\end{center}
\end{table*}

\begin{table*}[!htb] \tiny
	\begin{center}
		\vspace{-0mm}
		\caption{The performance of our proposed method with the comparison algorithms on 12 datasets. All algorithms are run on an RTX3090 GPU shader. }
		\vspace{-2mm}
		\label{T4}
		\resizebox{\linewidth}{!}{
			\begin{tabular}{c|c|cccccc}
				\toprule
				Dataset                             & Algorithm	   & Chebyshev $\downarrow$  & Clark $\downarrow$   &Canberra $\downarrow$   &K-L $\downarrow$    &Cosine $\uparrow$   &Intersection $\uparrow$           \\
				\midrule
				
				&  Ours        & 0.1239 $\pm$ \text{0.0091}           & 1.1699 $\pm$ \text{0.0108}       &2.1031  $\pm$ \text{0.0216}          & 0.1044 $\pm$ \text{0.0045}      & 0.9668 $\pm$ \text{0.0012}      & 0.8599 $\pm$ \text{0.0032}                 \\
				
				&  WUAKNN    & 0.1299 $\pm$ \text{0.0094}           & 1.1987 $\pm$ \text{0.0121}       &2.1280 $\pm$ \text{0.0432}          & 0.1101 $\pm$ \text{0.0034}     & 0.9612 $\pm$ \text{0.0023}      & 0.8446 $\pm$ \text{0.0066}                \\
				
				&  INP        & 0.1251 $\pm$ \text{0.0002}           & 1.1890 $\pm$ \text{0.0120}       &2.0980  $\pm$ \text{0.0223}          & 0.1053 $\pm$ \text{0.0009}      & 0.9643 $\pm$ \text{0.0015}      & 0.8501 $\pm$ \text{0.0025}                 \\
				
				&  LDL-LRR     & 0.1313 $\pm$ \text{0.0031}           & 1.2519 $\pm$ \text{0.0038}      &2.1992 $\pm$ \text{0.0095}          & 0.1127 $\pm$ \text{0.0077}     & 0.9533 $\pm$ \text{0.0021}      & 0.8412 $\pm$ \text{0.0066}                  \\
				
				&  LDL-LCLR    & 0.1277 $\pm$ \text{0.0016}          & 1.1969 $\pm$ \text{0.0039}       &2.1194 $\pm$ \text{0.0046}          & 0.1135 $\pm$ \text{0.0006}     & 0.9588 $\pm$ \text{0.0044}      & 0.8483 $\pm$ \text{0.0014}                  \\
				
				&  LDLSF       & 0.1270 $\pm$ \text{0.0028}           & 1.1909 $\pm$ \text{0.0164}       &2.1846 $\pm$ \text{0.0119}          & 0.1193 $\pm$ \text{0.0041}     & 0.9609 $\pm$ \text{0.0019}      & 0.8460 $\pm$ \text{0.0007}                     \\
				
				&  LALOT       & 0.1306 $\pm$ \text{0.0022}           & 1.1921 $\pm$ \text{0.0015}       &2.1111 $\pm$ \text{0.0171}          & 0.1120 $\pm$ \text{0.0015}     & 0.9430 $\pm$ \text{0.0019}      & 0.8400 $\pm$ \text{0.0004}                   \\
				
				\multirow{-7}{*}{fbp5500} 		            &  BFGS-LLD    & 0.1299 $\pm$ \text{0.0049}           & 1.4655 $\pm$ \text{0.0041}       &2.1675 $\pm$ \text{0.0024}          & 0.1135 $\pm$ \text{0.0055}     & 0.9595 $\pm$ \text{0.0030}      & 0.8419 $\pm$ \text{0.0018}                   \\
				
				\midrule
				
				&  Ours       & 0.1439 $\pm$ \text{0.0020}           & 1.3621 $\pm$ \text{0.0331}       &2.6799 $\pm$ \text{0.0045}          & 0.2011 $\pm$ \text{0.0006}      & 0.9434 $\pm$ \text{0.0044}      & 0.8329 $\pm$ \text{0.0095}                   \\
				
				&  WUAKNN     & 0.1499 $\pm$ \text{0.0053}           & 1.3996 $\pm$ \text{0.0432}       &2.7018 $\pm$ \text{0.0995}          & 0.2119 $\pm$ \text{0.0026}     & 0.9126 $\pm$ \text{0.0066}      & 0.8197 $\pm$ \text{0.0045}                \\
				
				&  INP        & 0.1456 $\pm$ \text{0.0021}           & 1.3651 $\pm$ \text{0.0441}       &2.6888 $\pm$ \text{0.0023}          & 0.2017 $\pm$ \text{0.0012}      & 0.9394 $\pm$ \text{0.0026}      & 0.8247 $\pm$ \text{0.0077}                   \\
				
				&  LDL-LRR     & 0.1526 $\pm$ \text{0.0033}           & 1.5651 $\pm$ \text{0.0111}      &2.7594 $\pm$ \text{0.0422}          & 0.2449 $\pm$ \text{0.0007}     & 0.9251 $\pm$ \text{0.0003}      & 0.8141 $\pm$ \text{0.0044}                      \\
				
				&  LDL-LCLR    & 0.1515 $\pm$ \text{0.0022}           & 1.592 $\pm$ \text{0.0117}       &2.7779 $\pm$ \text{0.0239}          & 0.2244 $\pm$ \text{0.0030}     & 0.9262 $\pm$ \text{0.0062}      & 0.8189 $\pm$ \text{0.0098}                   \\
				
				&  LDLSF       & 0.1488 $\pm$ \text{0.0024}           & 1.3889$\pm$ \text{0.0086}       &2.7672 $\pm$ \text{0.0660}          & 0.2302 $\pm$ \text{0.0044}     & 0.9111 $\pm$ \text{0.0051}      & 0.8117 $\pm$ \text{0.0022}                     \\
				
				&  LALOT       & 0.1479 $\pm$ \text{0.0010}           & 1.3659 $\pm$ \text{0.0099}       &2.6956 $\pm$ \text{0.0144}          & 0.2221 $\pm$ \text{0.0064}     & 0.9311 $\pm$ \text{0.0021}      & 0.8107 $\pm$ \text{0.0008}                    \\
				
				\multirow{-7}{*}{RAF-ML} 		            &  BFGS-LLD    & 0.1499 $\pm$ \text{0.0009}           & 1.6656$\pm$ \text{0.0066}       &2.7101 $\pm$ \text{0.0211}          & 0.2541 $\pm$ \text{0.0055}     & 0.9204 $\pm$ \text{0.0023}      & 0.8157 $\pm$ \text{0.0050}                    \\
				\midrule
				
				&  Ours        & 0.2763 $\pm$ \text{0.0088}           & 2.2313 $\pm$ \text{0.0114}       &5.1098 $\pm$ \text{0.0051}          & 0.5191 $\pm$ \text{0.0066}      & 0.8988 $\pm$ \text{0.0045}      & 0.7988 $\pm$ \text{0.0019}                  \\
				
				&  WUAKNN    & 0.2889 $\pm$ \text{ 0.0066}           & 2.3048 $\pm$ \text{0.0140}       &5.2136 $\pm$ \text{0.1556}          & 0.6077 $\pm$ \text{0.0049}     & 0.8657 $\pm$ \text{0.0066}      & 0.7764 $\pm$ \text{0.0033}                \\
				
				&  INP        & 0.2777 $\pm$ \text{0.0021}           & 2.2374 $\pm$ \text{0.0110}       &5.1163 $\pm$ \text{0.0018}          & 0.5111 $\pm$ \text{0.0029}      & 0.8807 $\pm$ \text{0.0049}      & 0.7891 $\pm$ \text{0.0014}                  \\
				
				&  LDL-LRR     & 0.3129 $\pm$ \text{0.0021}           & 3.2441 $\pm$ \text{0.0031}      &6.1454 $\pm$ \text{0.0023}          & 0.6616 $\pm$ \text{0.0035}     & 0.8002 $\pm$ \text{0.0042}      & 0.7411 $\pm$ \text{0.0014}                     \\
				
				&  LDL-LCLR    & 0.2994$\pm$ \text{0.0045}           & 2.4900 $\pm$ \text{0.0012}       &6.9609 $\pm$ \text{0.0041}          & 0.6056 $\pm$ \text{0.0031}     & 0.7110 $\pm$ \text{0.0021}      & 0.7110 $\pm$ \text{0.0088}                     \\
				
				&  LDLSF       & 0.3007 $\pm$ \text{0.0002}           & 2.7887 $\pm$ \text{0.0057}       &5.6101 $\pm$ \text{0.0118}         & 0.6396 $\pm$ \text{0.0022}     & 0.7939 $\pm$ \text{0.0098}      & 0.7660 $\pm$ \text{0.0007}                       \\
				
				&  LALOT       & 0.3133 $\pm$ \text{0.0021}           & 2.3141 $\pm$ \text{0.0016}       &5.5336 $\pm$ \text{0.0241}          & 0.5233 $\pm$ \text{0.0012}     & 0.8595 $\pm$ \text{0.0550}      & 0.7214 $\pm$ \text{0.0049}                    \\
				
				\multirow{-7}{*}{Twitter} 		            &  BFGS-LLD    & 0.3114 $\pm$ \text{0.0044}           & 2.5511 $\pm$ \text{0.0028}       &5.7145 $\pm$ \text{0.0041}          & 0.5461 $\pm$ \text{0.0153}     & 0.8335 $\pm$ \text{0.0055}      & 0.7744 $\pm$ \text{0.0020}     \\
				
				\midrule
				
				&  Ours        & 0.2821 $\pm$ \text{0.0019}           & 2.3155 $\pm$ \text{0.0066}       &5.2189 $\pm$ \text{0.0099}          & 0.5215 $\pm$ \text{0.0035}      & 0.8413 $\pm$ \text{0.0045}      & 0.7803 $\pm$ \text{0.0026}                 \\
				
				&  WUAKNN     & 0.3210 $\pm$ \text{0.0025}           & 2.6655 $\pm$ \text{0.1003}       &5.5610 $\pm$ \text{0.0033}          & 0.6045 $\pm$ \text{0.0099}     & 0.8321$\pm$ \text{0.0138}      & 0.7650 $\pm$ \text{0.0022}                \\
				
				&  INP        & 0.2816 $\pm$ \text{0.0031}           & 2.3356 $\pm$ \text{0.0097}       &5.2222 $\pm$ \text{0.0159}          & 0.5314 $\pm$ \text{0.0033}      & 0.8406 $\pm$ \text{0.0041}      & 0.7741 $\pm$ \text{0.0025}                 \\
				
				&  LDL-LRR     & 0.3329 $\pm$ \text{0.0012}           & 3.4400 $\pm$ \text{0.0174}      &6.3459 $\pm$ \text{0.0229}          & 0.6516 $\pm$ \text{0.0031}     & 0.8450$\pm$ \text{0.0040}      & 0.7399 $\pm$ \text{0.0037}                    \\
				
				&  LDL-LCLR    & 0.2970$\pm$ \text{0.0009}           & 2.4444 $\pm$ \text{0.0063}       &6.1600 $\pm$ \text{0.0041}         & 0.6222 $\pm$ \text{0.0013}     & 0.7919 $\pm$ \text{0.0029}      & 0.7090$\pm$ \text{0.0070}                     \\
				
				&  LDLSF       & 0.3301 $\pm$ \text{0.0009}           & 2.8888 $\pm$ \text{0.0459}       &5.9152 $\pm$ \text{0.0121}          & 0.6100 $\pm$ \text{0.0021}     & 0.8139 $\pm$ \text{0.0098}      & 0.7360 $\pm$ \text{0.0037}                      \\
				
				&  LALOT       & 0.3411 $\pm$ \text{0.0026}           & 2.9140 $\pm$ \text{0.0019}       &5.3333 $\pm$ \text{0.0243}          & 0.5737 $\pm$ \text{0.0012}     & 0.8225 $\pm$ \text{0.0202}     & 0.7144 $\pm$ \text{0.0004}                    \\
				
				\multirow{-7}{*}{Flickr} 		            &  BFGS-LLD    & 0.3200 $\pm$ \text{0.0041}           & 2.7517 $\pm$ \text{0.0060}       &5.8149 $\pm$ \text{0.0048}          & 0.5961 $\pm$ \text{0.0099}     & 0.8131 $\pm$ \text{0.0011}      & 0.7407 $\pm$ \text{0.0077}       \\
				
				\bottomrule
			\end{tabular}%
		}
		\vspace{-4mm}
	\end{center}
\end{table*}
\begin{table*}[!htb] \scriptsize
	\begin{center}
		\vspace{-2mm}
		\caption{Ablation study. Effectiveness of the loss functions and the modules on \textit{Gene}. Quantitative results demonstrate the effectiveness of each module.}
		\vspace{-2mm}
		\label{T5}
		\resizebox{\linewidth}{!}{
			\begin{tabular}{c|cccccc}
				\toprule
				Algorithm	   & Chebyshev $\downarrow$  & Clark $\downarrow$   &Canberra $\downarrow$   &K-L $\downarrow$    &Cosine $\uparrow$   &Intersection $\uparrow$                \\

				\midrule
				Ours        & 0.0481 $\pm$ \text{0.0036}           & 2.1010 $\pm$ \text{0.0256}       &14.0802 $\pm$ \text{0.0154}          & 0.2333 $\pm$ \text{0.0091}      & 0.8409 $\pm$ \text{0.0030}      & 0.7993 $\pm$ \text{0.0022}                   \\
				
				w/o prototype matching       & 0.0502 $\pm$ \text{0.0013}           & 2.1531 $\pm$ \text{0.0111}       &14.1366 $\pm$ \text{0.0101}          & 0.2355 $\pm$ \text{0.0052}      & 0.8329 $\pm$ \text{0.0031}      & 0.7893 $\pm$ \text{0.0021}                   \\
				
				w/o micro-perturbation   & 0.0485 $\pm$ \text{0.0012}           & 2.1019 $\pm$ \text{0.0133}       &14.0812 $\pm$ \text{0.0033}          & 0.2349 $\pm$ \text{0.0112}      & 0.8398 $\pm$ \text{0.0033}      & 0.7990 $\pm$ \text{0.0025}                   \\

				\bottomrule
			\end{tabular}%
		}
		\vspace{-4mm}
	\end{center}
\end{table*}

\begin{figure*}[!htb]\scriptsize
	\begin{center}
		\vspace{-2mm}
		\tabcolsep 1pt
		\begin{tabular}{@{}cccc@{}}
			\includegraphics[width = 0.24\textwidth]{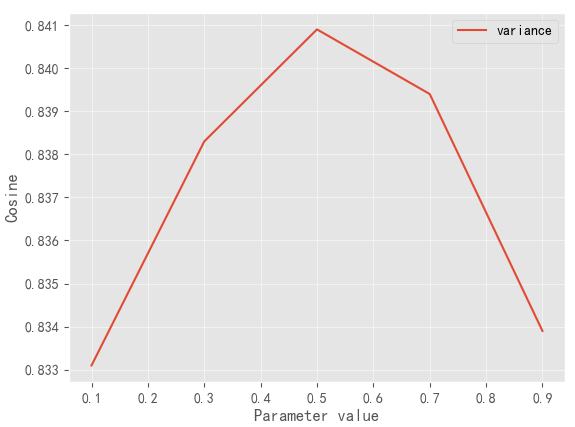} &
			\includegraphics[width = 0.24\textwidth]{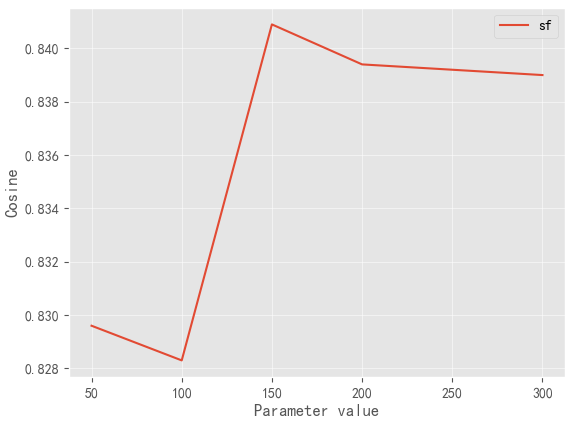}       & 
			\includegraphics[width = 0.24\textwidth]{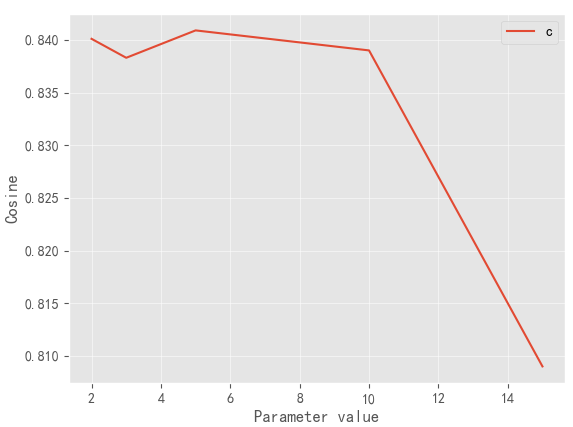}       & 
			\includegraphics[width = 0.24\textwidth]{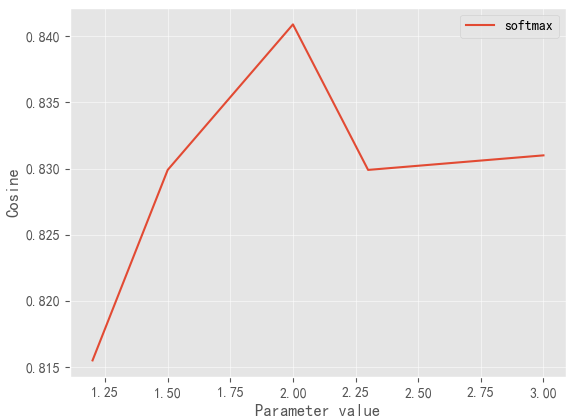}   \\
			(a) variance &
			(b) $sf$ & 
			(c) $c_{i}$ &
			(d) $\text{Softmax}^{*}$
			\\
		\end{tabular}
	\end{center}
	\vspace{-6mm}
	\caption{This figure shows the sensitivity of parameters on the Gene dataset.}
	\vspace{-4mm}
	\label{g}
\end{figure*}

\noindent \textbf{Parameter Sensitivity Analysis.}
Our method has four parameters, including the variance of the Gaussian function, the sampling frequency, the proportion of $c_{i}$ occupied in $\mu_{i}$, and a real number in $\text{Softmax}^{*}$ replaces the index e.
To analyze the sensitivity of variance, the sampling frequency $sf$, the proportion of $c_{i}$ occupied in $\mu_{i}$, and $\text{Softmax}^{*}$, we run our method with four sets (\{0.1, 0.3, 0.5, 0.7, 0.9\}, \{50, 100, 150, 200, 300\}, \{2\%, 3\%, 5\%, 10\%, 15\%\} and \{1.2\%, 1.5\%, 1.8\%, 2.1\%, 2.4\%\}) on the Gene dataset (see Figure~\ref{g}). We primarily evaluate the metric of cosine distance.

\begin{table*}[!htb] \tiny
	\begin{center}
		\vspace{-0mm}
		\caption{The performance of our proposed method with the comparison algorithms on the Gene dataset (a new assessment preprocessing technique is used). All algorithms are run on an RTX3090 GPU shader. }
		\vspace{-2mm}
		\label{T9}
		\resizebox{\linewidth}{!}{
			\begin{tabular}{c|c|cccccc}
				\toprule
				Dataset                             & Algorithm	   & Chebyshev $\downarrow$  & Clark $\downarrow$   &Canberra $\downarrow$   &K-L $\downarrow$    &Cosine $\uparrow$   &Intersection $\uparrow$           \\
				\midrule
				&  Ours             & 0.0686 $\pm$ \text{0.0012}      & 0.4899 $\pm$ \text{0.0100}       & 0.7522 $\pm$ \text{0.0133}      & 0.3996 $\pm$ \text{0.0022}       & 0.9550 $\pm$ \text{0.0013}      & 0.8779 $\pm$ \text{0.0003}                   \\
				
				&  WUAKNN     & 0.0732 $\pm$ \text{0.0033}      & 0.5112 $\pm$ \text{0.0202}       & 0.7888$\pm$ \text{0.0030}      & 0.4556 $\pm$ \text{0.0111}       & 0.9230 $\pm$ \text{0.0013}      & 0.8653 $\pm$ \text{0.0005}                   \\
				
				&  INP        & 0.0690 $\pm$ \text{0.0093}      & 0.4902$\pm$ \text{0.0103}       & 0.7609 $\pm$ \text{0.0099}      & 0.4002 $\pm$ \text{0.0023}       & 0.9441 $\pm$ \text{0.0006}      & 0.8667 $\pm$ \text{0.0005}                   \\
				
				&  LDL-LRR     & 0.0711 $\pm$ \text{0.0043}      & 0.5222 $\pm$ \text{0.0066}       & 0.7778 $\pm$ \text{0.0022}      & 0.4463 $\pm$ \text{0.0033}       & 0.9301 $\pm$ \text{0.0005}      & 0.8554 $\pm$ \text{0.0019}                   \\
				
				&  LDL-LCLR    & 0.0898 $\pm$ \text{0.0066}      & 0.5213 $\pm$ \text{0.0050}       & 0.7965 $\pm$ \text{0.0286}      & 0.5006 $\pm$ \text{0.0044}       & 0.9119 $\pm$ \text{0.0099}      & 0.8336 $\pm$ \text{0.0066}                   \\
				
				&  LDLSF       & 0.0750 $\pm$ \text{0.0023}      & 0.5665 $\pm$ \text{0.0033}       & 0.7653 $\pm$ \text{0.0043}      & 0.4009 $\pm$ \text{0.0007}       & 0.9333 $\pm$ \text{0.0020}      & 0.8565 $\pm$ \text{0.0080}                   \\
				
				&  LALOT       & 0.0702 $\pm$ \text{0.0203}      & 0.4995 $\pm$ \text{0.0098}       & 0.7603 $\pm$ \text{0.0045}      & 0.4008 $\pm$ \text{0.0003}       & 0.9508 $\pm$ \text{0.0033}      & 0.8557 $\pm$ \text{0.0077}                   \\
				
				\multirow{-7}{*}{Gene} 		&  BFGS-LLD              & 0.8897 $\pm$ \text{0.0032}      & 0.5880 $\pm$ \text{0.0111}       & 0.8095 $\pm$ \text{0.0022}      & 0.4440 $\pm$ \text{0.0135}       & 0.9023 $\pm$ \text{0.0043}      & 0.8779 $\pm$ \text{0.0033}                                       \\
				
				\bottomrule
			\end{tabular}%
		}
		\vspace{-4mm}
	\end{center}
\end{table*}
\begin{table*}[!htb] \tiny
	\begin{center}
		\vspace{-0mm}
		\caption{The performance of our proposed method with the comparison algorithms on the Gene dataset (ensemble learning is used). All algorithms are run on an RTX3090 GPU shader. }
		\vspace{-2mm}
		\label{T10}
		\resizebox{\linewidth}{!}{
			\begin{tabular}{c|c|cccccc}
				\toprule
				Dataset                             & Algorithm	   & Chebyshev $\downarrow$  & Clark $\downarrow$   &Canberra $\downarrow$   &K-L $\downarrow$    &Cosine $\uparrow$   &Intersection $\uparrow$           \\
				\midrule
				&  Ours        & 0.0466 $\pm$ \text{0.0062}           & 2.0930 $\pm$ \text{0.0111}       &12.1888 $\pm$ \text{0.0112}          & 0.1887 $\pm$ \text{0.0009}      & 0.8508 $\pm$ \text{0.0031}      & 0.8007 $\pm$ \text{0.0021}                   \\
				
				&  WUAKNN     & 0.0511 $\pm$ \text{0.0021}           & 2.1899 $\pm$ \text{0.0016}       &13.8556 $\pm$ \text{0.1913}          & 0.2368 $\pm$ \text{0.0121}     & 0.8409 $\pm$ \text{0.0013}      & 0.7820 $\pm$ \text{0.0020}                \\
				
				&  INP        & 0.0480 $\pm$ \text{0.0012}           & 2.0998 $\pm$ \text{0.0159}       &14.0144 $\pm$ \text{0.0120}          & 0.2309 $\pm$ \text{0.0045}      & 0.8396 $\pm$ \text{0.0066}      & 0.8003 $\pm$ \text{0.0013}                   \\
				
				&  LDL-LRR     & 0.0511 $\pm$ \text{0.0004}           & 2.1997 $\pm$ \text{0.0860}      &13.9660 $\pm$ \text{0.0143}          & 0.2489 $\pm$ \text{0.0036}     & 0.8399 $\pm$ \text{0.0111}      & 0.7895$\pm$ \text{0.0044}                     \\
				
				&  LDL-LCLR    & 0.0509 $\pm$ \text{0.0044}           & 2.1339 $\pm$ \text{0.0088}       &13.7789 $\pm$ \text{0.0065}          & 0.2499$\pm$ \text{0.00199}     & 0.8408 $\pm$ \text{0.0087}      & 0.7698$\pm$ \text{0.0066}                    \\
				
				&  LDLSF       & 0.0505 $\pm$ \text{0.0025}           & 2.1881$\pm$ \text{0.0033}       &13.7965 $\pm$ \text{0.0199}          & 0.2333 $\pm$ \text{0.0018}     & 0.8408 $\pm$ \text{0.0088}      & 0.7756 $\pm$ \text{0.0032}                      \\
				
				&  LALOT       & 0.0498 $\pm$ \text{0.0034}           & 2.1565 $\pm$ \text{0.0087}       &13.8989 $\pm$ \text{0.0156}          & 0.2398 $\pm$ \text{0.0002}     & 0.8300 $\pm$ \text{0.0051}      & 0.7895 $\pm$ \text{0.0035}                     \\
				
				\multirow{-7}{*}{Gene} 		            &  BFGS-LLD    & 0.0566$\pm$ \text{0.0036}           & 2.3209 $\pm$ \text{0.0099}       &14.1665 $\pm$ \text{0.0998}          & 0.2598 $\pm$ \text{0.0014}     & 0.8302$\pm$ \text{0.0044}      & 0.7789 $\pm$ \text{0.0031}                     \\
				\bottomrule
			\end{tabular}%
		}
		\vspace{-4mm}
	\end{center}
\end{table*}

\noindent \textbf{Results and analysis.}
We conduct 10 times 5-fold cross-validation on each dataset (note that we use a hierarchical (cluster) approach to statistically construct the training and testing samples). 
The experimental results are presented in the form of ``mean$\pm$std'' in Tables~\ref{T3} and ~\ref{T4}.
Overall, our proposed method outperforms other comparison algorithms on all evaluation metrics.
We deduce several interesting insights from the report.
\textbf{i):} From the experimental results, the top three algorithms in terms of performance are UAKNN, INP, and WUAKNN. Both our method and INP take into account the uncertainty of the label space, which may be the reason why the algorithm is so competitive. 
%
%
\textbf{ii):} Although INP closely follows our approach in terms of performance, it requires an elaborate set of networks and parameters for each dataset, which is extremely costly.
\textbf{iii):} Cross-validation with randomness inevitably introduces the out-of-distribution problem. In brief, the distributions of the training and test samples are not \textit{i.i.d}.
The KNN-type approach seems to work for it because it avoids overtraining a set of parameters matching the distribution of the training set.
Moreover, we evaluate the range of p-values for the six metrics on 12 data sets.
Cheby.$[1.22e-103, 1.00e+01]$, Clark$[5.66e-94, 1.14e-03]$, Canbe.$[10.12e-99, 1.12e-01]$, KL$[1.53e-103, 1.23e-03]$, Cosine$[1.35e-99, 1.21e-02]$, and Inter.$[1.41e-110, 7.85e-02]$
According to the test results, the LDL methods have significantly different performances in terms of each metric on all datasets except \textit{Gene} (at a 0.05 significance level). 
%

\noindent \textbf{Ablation study.}
To demonstrate the effectiveness of  the module of our model, we conduct an ablation study involving the following two experiments:
\textbf{(a)} w/o prototype matching: We run the search on the whole training dataset and avoid dividing the prototypes (clusters), shown in Table~\ref{T5}.
\textbf{(b)} w/o micro-perturbation: We remove the micro-perturbation $c_{i}$ in the assignment of the weight, shown in Table~\ref{T5}.
Overall, the prototype (cluster) matching approach has a greater benefit to the overall algorithm enhancement, and the micro-perturbation boosts the generalization ability of the model.
For the model's inference speed, the removed operations bring less than 2\% speedup.
This experiment is conducted with a 5-fold cross-validation.
\vspace{-4mm}
\section{Specialized for Gene datasets}
\vspace{-2mm}
In this section, we attempt to model two new strategies to specialize for \textit{Gene} datasets.
The \textit{Gene} dataset contains 68 labels,  and none of the extant LDL algorithms struggle to distinguish their model capabilities on the 6 customized metrics.
The reasons may come from two aspects, 1) It is an extreme label problem in LDL tasks; 2) The sum of all the labels is \textbf{1} which limits the distinguishability of the labels in the large label space.
To alleviate this problem, we introduce a new label normalization scheme that acts on the label space, and an ensemble learning framework to boost the modeling capability of existing methods.
\vspace{-2mm}

\noindent \textbf{A new label normalization scheme.}
%
%
In real-world scenarios, researchers or engineers focus on a few key labels rather than dealing with all outputs in an omnibus fashion.
\textit{Gene} dataset that contains 68 outputs may not all be critical, and in addition, energizing the thin label signal boosts variance.
Based on the above-mentioned philosophy, we design a label normalization scheme to replace Softmax.
Specifically, the first step is to revise the label format of our test samples.
On the label distribution $\mathcal{D}_{i}$ (treated as an array) to retrieve label order numbers $LON$ with label values greater than 0.014.
Next, the label values are stitched together into a label vector $\mathcal{D}^{*}_{i}$ and normalized by $\text{Softmax}^{*}$.
Finally, the predicted label distribution $\mathcal{L}_{i}$ is stitched back together by $LON$ and $\text{Softmax}^{*}$ into a new predicted label distribution $\mathcal{L}^{*}_{i}$ for comparison with $\mathcal{D}^{*}_{i}$ (see Figure~\ref{T9}).
Comparing $\mathcal{L}^{*}_{i}$ and $\mathcal{D}^{*}_{i}$ still use the six customized metrics.
Note that the purpose of this label regularization method is to distinguish the performance of different algorithms on the \textit{Gene} dataset.
In brief, the ``energy'' (description degree) of important labels is enhanced.

\vspace{-2mm}

\noindent \textbf{Ensemble learning framework.}
Besides changing the metrics for evaluating the performance of the model, we observe that the feature space of \textit{Gene} is less than the label space, which is a classic extreme learning problem (in other words, this is an ill-posed problem).
Ensemble learning finishes excellent in tackling the extreme learning problem~\cite{liu2010ensemble}, we attempt to build a model ensemble (including 5 basic models, and the predicted results are averaged) for each LDL algorithm by employing different parameters on the training set.
The experimental results are shown in Figure~\ref{T10}, where the performance of all methods on the \textit{Gene} dataset is boosted while the variance of the performance comparison is amplified.
In addition, the differences between algorithms are amplified along with the increase in the number of ensemble models.

\vspace{-5mm}

\section{Limitations}
\vspace{-3mm}
Although UAKNN can be conveniently adapted to arbitrary datasets (the LDL dataset has feature dimensions from 36 to 1869 and label dimensions from 5 to 68 on 12 benchmarks), its decision-making capability depends on the modeling of pre-processing models.
As an example, \textit{wc-LDL} is an image dataset whose image resolution is uniformly cropped to $256 \times 256$.
We employ three pre-trained models (ResNet18, VGG19, ResNet50) to project an image to a vector of dimension 243, respectively.
For ResNet18, our method is $0.9899 \pm 0.0013$ on the \textit{Cosine} metric.
For VGG19 and ResNet50, our method obtains $0.9870 \pm 0.0022$ and $0.9909 \pm 0.0005$, respectively.
For the rest of the LDL algorithms, the difference in the ability of the pre-trained models to impact the downstream tasks is not significant.
Besides, UAKNN is tensorized with the help of \texttt{skorch}, and its inference speed is accelerated on the GPU shader.
However, UAKNN runs significantly slower on the CPU.
For example, on the \textit{Movie} dataset, UAKNN spends 16ms to run a test sample for inference, while on the CPU it requires nearly 40ms.
Notably, some of its optimized properties come from the benefits of PyTorch 2.0.
\vspace{-5mm}
\section{Conclusion}
\vspace{-3mm}
This paper shows a novel insight that models without parameters can be competitive on LDL tasks.
Extensive experiments demonstrate that the low-rank characteristic, uncertainty, and micro-perturbation boost the modeling ability of the UAKNN.
In addition, we also propose some strategies that are useful for modeling extremely label distribution datasets.
Overall, UAKNN has accurate modeling capabilities and real-time inference and can be rapidly deployed without training on arbitrary LDL datasets.
\clearpage

\bibliography{uai2023-template}
\end{document}